%% file: main.tex
\documentclass[a4paper,12pt]{article}
\usepackage[utf8]{inputenc} 
\usepackage[final]{nejlt}
\usepackage{fullpage}
\usepackage{covington} 
\usepackage{natbib}
\usepackage{graphicx}
\usepackage{multicol}
\usepackage{pgfplots}
\usepackage{booktabs}
\usepackage{tabularx}
\usepackage{longtable}
\usepackage{lscape}
\usepackage{subcaption}
\usepackage{pgfplotstable}
\usepackage{tikz}
\usetikzlibrary{external,fit,positioning}
\usepackage[section]{placeins}
\usepackage{xurl}

\newcommand{\DEVELOPMENT}{0} 
\usepackage{ifthen}
\ifthenelse{\DEVELOPMENT = 1}{
	\newcommand{\da}[1]{\textcolor{cyan}{\textbf{DA:} #1}}		
	\newcommand{\ev}[1]{\textcolor{magenta}{\textbf{EV:} #1}}	
    \newcommand{\tlt}[1]{\textcolor{purple}{\textbf{TLT:} #1}}
}{
	\newcommand{\da}[1]{}		
	\newcommand{\ev}[1]{}
    \newcommand{\tlt}[1]{}
}

\newcommand{\ANON}{0}

\ifthenelse{\ANON = 1}{
\newcommand{\anonurld}[1]{\textsc{Anonymized URL}}
\newcommand{\anon}[1]{\textsc{Anonymized}}
\newcommand{\neveranonurl}[1]{#1}
\let\url=\anonurld
\newcommand{\anoncitep}[1]{\textsc{(Anonymized citation)}}
\newcommand{\anoncitet}[1]{\textsc{Anonymized citation}}
}{
\newcommand{\anon}[1]{#1}
\newcommand{\neveranonurl}[1]{#1}
\newcommand{\anoncitep}[1]{\citep{#1}}
\newcommand{\anoncitet}[1]{\citet{#1}}
}

\title{Crowdsourcing Relative Rankings of Multi-Word\\Expressions:
Experts versus Non-Experts
}

\author{David Alfter$^{1}$, Therese Lindström Tiedemann$^{2}$, Elena Volodina$^{1}$\\ 
 [0.5cm] $^{1}$University of Gothenburg \hspace{12mm} $^{2}$University of Helsinki\\ 
 Spr\aa kbanken \hspace{40mm} Department of Finnish, Finno-Ugrian \\
 Department of Swedish \hspace{18.5mm} and Scandinavian Studies\\ 
[0.5cm] \texttt{david.alfter@gu.se}\\\texttt{therese.lindstromtiedemann@helsinki.fi}\\\texttt{elena.volodina@gu.se}}   

\date{}

\pgfplotsset{compat=1.16}
\begin{document}
\thispagestyle{empty}

\abstract{
In this study we investigate to which degree experts and non-experts agree on questions of difficulty in a crowdsourcing experiment. We ask non-experts (second language learners of Swedish) and two groups of experts (teachers of Swedish as a second/foreign language and CEFR experts) to rank multi-word expressions in a crowdsourcing experiment. We find that the resulting rankings by all the three tested groups correlate to a very high degree, which suggests that judgments produced in a comparative setting are not influenced by  
professional insights into Swedish as a second language. 
}

\maketitle

\section{Introduction}
\input{sections/1_introduction.tex}
\label{section:introduction}

\section{Related Work}
\input{sections/2_related_work.tex}
\label{section:related:work}

\section{Data}
\input{sections/3_data.tex}
\label{section:data}

\section{Methodology}
\input{sections/4_methodology.tex}
\label{section:methodology}

\section{Experimental Setup}

\input{sections/5_0_experimental_setup.tex}

\label{section:experimental:setup}

\section{Results and Analysis}
\input{sections/6_0_results.tex}
\label{section:results:debug}

\section{Discussion}
\input{sections/7_discussion.tex}
\label{section:discussion}

\section{Conclusion}
\input{sections/8_conclusion.tex}
\label{section:conclusion}

\bibliographystyle{nejlt_bib}
\bibliography{bibliography}
\end{document}

%% file: sections/1_introduction.tex
Many of the challenges in automatically driven solutions for language learning boil down to the lack of data and resources based on which we can develop language learning materials or train models. Resources like the English Vocabulary Profile \cite{capel2010a1,capel2012completing,evp} are a luxury that cost a lot of time and resources to create, and for most languages such resources do not exist. Crowdsourcing has been suggested 
as one of the potential methods to overcome these challenges. Recently, a European network enet-Collect\footnote{\neveranonurl{https://enetcollect.eurac.edu/}} \cite{lyding2018introducing} has been initiated to stimulate synergies between language learning research and practice on the one hand, and crowdsourcing on the other. New initiatives have arisen as a result, e.g. using implicitly crowdsourced learner knowledge for language resource creation \cite{nicolas2020creating}, crowdsourcing corpus cleaning \cite{kuhn2019crowdsourcing}, development of the Learning and Reading Assistant LARA \cite{habibi2019lara}.  However, there are many questions that need to be investigated and answered with regards to methodological issues arising from using crowdsourcing as a method in/for Language Learning.  


In this article, we raise some methodological questions about crowdsourcing in the context of second language (L2) learning material creation. To go back to the example of the English Vocabulary Profile -- could we generate something similar for other languages without involving lexicographers and experts? For example, given a set of some unordered vocabulary items (e.g. phrases), how can we order them by difficulty and split them into groups appropriate for teaching at different levels of linguistic proficiency? Could a crowd help us in this scenario? Who can be ``the crowd" in that case? How many answers are enough? How many contributors are needed? Are the results reliable? Parts of this article have been described in \newcite{alfter2020expert}. 

We focus on whether a crowd of non-expert crowdsourcers 
can be used to generate language learning materials
and how the annotations by experts such as L2 Swedish teachers, assessors and researchers, i.e.\ people with formal training in teaching and assessing in Swedish, 
compare to the annotations by non-experts, by which we mean learners and speakers of L2 Swedish.\footnote{By L2 Swedish we mean Swedish as a second (third, fourth, \ldots) language and as a foreign language} 
On a more general note, we investigate whether crowdsourcing as a method can be \emph{reliably} applied to language learning resource building using a mixed crowd.

We use a selection of multi-word expressions (MWE) and ask experts (teachers, assessors etc) and non-experts (language learners) to arrange MWEs by difficulty. The crowdsourcing part of the experiment is designed in such a way that we test which \emph {intuitions} people have about the relative difficulty of \emph{understanding}
word combinations. In this design, we do not expect our participants to know anything explicitly about language learning theories, instead relying on their intuitive comparative judgments as intuitive comparative judgments -- including ranking items against each other -- has been proven to be easier than assigning items to a category (e.g. a level of proficiency) \cite{lesterhuis2017comparative}.
We hypothesize that given an unordered list of expressions, using crowdsourcing, we can derive a list ordered by  difficulty that can be used in language teaching.
We surmise that difficulty and proficiency are correlated, thus one might expect more difficult expressions to be learned at later stages of language development. 

The theoretical
notion of \textit{L2 proficiency} is of special importance in connection to this study.
Proficiency is a key concept in Second Language Acquisition (SLA) research. It is used to describe the language ``knowledge, competence, or ability" 
(\newcite[p.~16]{bachman1990fundamental}, as cited in 
\newcite[p.~163]{carlsen2012proficiency}) 
of a learner on a conventionalized scale, one example being the 6-level scale adopted by the Common European Framework of Reference (CEFR) \cite{councilofeurope2001,coe2018common}.  Conventionalized scales of proficiency levels are useful in educational and assessing contexts, e.g. which group to place a student into \cite{bachman2010language} and in various social and political scenarios, e.g. whether an applicant can be granted citizenship \cite{forsberglundell2020krav}. However, a straightforward division into levels is a tricky endeavor, since there is no consensus how to define a level and its corresponding competence(s) in concrete terms. SLA research is specific about viewing proficiency as a ``coarse-grained, externally motivated" construct \cite[p.~134]{ortega2012interlanguage}, where levels are always somewhat arbitrary \cite[p.~17]{councilofeurope2001} and proficiency should be seen as different to \textit{L2 development} which is ``an internally motivated trajectory of linguistic acquisition" \cite[p.~134]{ortega2012interlanguage}. 

For this reason, current approaches to proficiency advocate rather a scalar/interval approach as it is more powerful, realistic and nuanced \cite{ortega2012interlanguage,coe2018common,paquot2020using}. The current experiment is proof of the usefulness of such an approach where rather than stating that certain vocabulary belongs to a certain level, we can instead state that some vocabulary items are perceived as easier or more difficult in comparison to each other and form a growing scale of items which are likely to be learned in that approximate order.




This article is structured as follows: 
we introduce related work in Section 2 and describe the data used for the experiment is Section 3.  Sections 4 and 5  introduce Methodology and Experimental design. In Sections 6 and 7 we present our main results, analyze and discuss them. Section 8 concludes the article. 



%% file: sections/2_related_work.tex
Previously, several approaches have been used in identifying and ordering relevant vocabulary items for second language learning.
A popular approach is to use reading material written by first language (L1) speakers such as newspapers to generate frequency-based word lists, e.g. the Kelly lists \cite{kilgarriff2014corpus}, the General Service List \cite{west1953general}, the New GSL \cite{brezina2015there}.
Such word lists tend to use frequency of occurrence in a corpus as the only criterion for deciding which items 
that should be taught first, and which ones should be introduced later, following the hypothesis that more frequent words would be easier (cf e.g. \newcite{eskildsen2009constructing} regarding usage-based approaches to L2 acquisition)
and more important to know, while more rare words would be more difficult and less critical for communication in a target language. 
While such lists are useful, they also have drawbacks, especially in the context of second language learning. Indeed, L1 reading material is rarely adequate for language learner needs and lacks important vocabulary items \cite[p.~3767]{franccois2014flelex}.

In order to address the L2 learner needs, there has also been work on using L2 materials as a basis for word lists.
One possible approach is to use graded textbooks as a starting point, as has been done in the CEFRLex project.\footnote{\neveranonurl{https://cental.uclouvain.be/cefrlex/}} The motivation behind this approach is that textbooks generally target a specific proficiency group of language learners and have been carefully written with the needs of second language learners in mind. The project so far has resulted in the creation of six corpus-based language lists in six languages: FLELex for French \cite{franccois2014flelex}, SVALex for Swedish \cite{franccois2016svalex}, EFLLex for English \cite{durlich2018efllex}, NT2Lex for Dutch \cite{tack2018nt2lex} and ELELex for Spanish \cite{franccois2018elelex}. Each of these word lists not only contains the overall frequency but also the distribution of frequencies over the different CEFR levels. These projects have assumed that, in theory, the level at which a text is used in a language learning scenario can be used as an indication of a level at which vocabulary of that text can be assumed to be understood by learners and thus can be qualified as a learning target. In practice, however, this relationship is not as straightforward (e.g. \newcite{benigno2019linking}). 

Another approach based on L2 material is to use graded learner essays. This has been done in projects such as the English Vocabulary Profile (EVP)\footnote{\neveranonurl{https://www.englishprofile.org/wordlists}} \cite{capel2010a1,capel2012completing} and SweLLex \cite{volodina2016swellex}. SweLLex belongs to the CEFRLex family, as it has been created with the methodology behind CEFRLex, but in contrast to other resources in the family, it is based on learner essays, more specifically the SweLL pilot corpus \cite{volodina2016swell} of graded essays written by learners of Swedish. Both of these resources have also experimented with a threshold approach to assigning levels \cite{hawkins2012criterial,alfter2016distributions}, i.e. taking as indicative level not simply the first occurrence but the first \emph{significant} occurrence, i.e. the level at which a word or expression is used a certain number of times as defined by a threshold value. 
Deriving word lists from learner essays may prove more reliable as the amount of data increases \cite{pilan2016predicting}, and when the non-standard learner language has been effectively standardized (i.e. corrected) 
to the target language forms since automatic annotation is almost always trained on standard L1 materials  \cite[cf.][]{stemle2019working}. Both aspects, however, are non-trivial and very few languages enjoy the luxury of extensive corrected 
collections of learner-produced data. 

Finally, one can consult L2 experts 
to rely on their judgments as to the difficulty of items. 
Expert judgment as a method has been widely applied in general linguistics as well as in second language oriented experiments and L2 resource creation \cite[e.g.][]{spinner2019judgment,capel2010a1,capel2012completing}, 
although not without criticism. One of the potential stumbling blocks is the \emph{subjective intuitive} nature of judgments, something which is claimed to be a major obstacle to reliable scientific conclusions; observations, i.e. language \emph{production}, are regarded as a more reliable and desirable source of data \cite{bloomfield1935linguistic}.
However, \newcite{chomsky1965some} 
argue that judgments versus observations reflects the dichotomy between competence versus performance.
In the end, expert judgments reflect experts' professional experience, and are based on evidence coming from \textit{their} practices and theoretical assumptions about L2 teaching, and thus inevitably reflect personal interpretations of these. The challenge is, thus, to overcome the subjectivity of judgments without losing correctness of the final conclusions, so that the results can be used as a basis for assumptions about language learning paths and for scheduling learning materials in an optimal (although obviously never perfect) way. 
By \textit{direct labeling} we mean that experts explicitly label each item with a CEFR level (A1-C2+). This method is also  referred to as the \textit{``Hey Sally"} method in \newcite{spinner2019judgment}, indicating decision making based on consulting with other expert colleagues to either reduce or confirm the personal subjective bias. 

Due to problems with the reliability of manual level assignment, some people have experimented with the number of experts and procedures that would be necessary to gain reliable objective results. \newcite{carlsen2012proficiency} notes that the Norwegian L2 corpus project ASK \cite{tenfjord2006ask} used 10 CEFR assessors for their essays, who for the most part worked in groups of 5 so that each essay was marked by at least 5 assessors to get a reliable result. Similarly,  \newcite{lenko2015english} used 2-4 raters for the level assignment of her subset of the international corpus of learner English (ICLE) \cite{granger2009international} to reach agreement between the raters and \newcite{diez2012use} reported very low inter-rater reliability when using only 2 raters to assign CEFR-levels to Spanish university entrance exams. 
Furthermore, previous research has shown that the background of the rater is of much importance (cf \newcite{diez2012use} for an overview), although the results have been mixed. Experienced raters have sometimes rated more strictly (\newcite{sweedler1985influence} as cited in \newcite{diez2012use}) but in other studies they were more lenient (\newcite{weigle1998using} as cited in \newcite{diez2012use}). Whether the rater is a native speaker or not has also been seen to have an effect, in addition to gender, but once again the results were mixed. \newcite{diez2012use} also shows that how different rater backgrounds rate the proficiency has also depended on whether holistic or analytic scales were used.

The inherent order of teaching the items on the various vocabulary lists 
mentioned earlier, however, is not always obvious. 
Frequencies can be misleading, insufficient or sometimes idiosyncratic. 
Expert judgments might be perceived as less idiosyncratic 
but can be inaccessible due to the costs entailed in expert work. 
Crowdsourcing as a method of annotation could be worth exploring to address the above mentioned weaknesses. 

To the best of our knowledge, crowdsourcing has not been extensively used for such ordering tasks. However, we surmise that it might be an alternative to the more heavily resource reliant methods.
Crowdsourcing can take different forms. On the one hand, it can be quite explicit 
about the crowdsourcing aspect. In its original form,
it would consist in the annotation of the same data by different annotators \cite{fort2016collaborative} or the collaborative creation and curation of resources such as Wikipedia \cite{stegbauer2009wikipedia}. Such forms generally rely on intrinsic motivation. However, if there is a lack of intrinsic motivation for whatever reasons, two different approaches have been taken, the first of which is paying people, and the second of which is making the task more fun by adding game-like elements \cite{chamberlain2013using}. The monetary aspect is expressed in platforms such as Amazon Mechanical Turk which pays participants to answer questions and/or solve tasks \cite{buhrmester2016amazon}. 
On the other hand, crowdsourcing can be more subtle, such as in Games With A Purpose (GWAPS) \cite{lafourcade2015games}. GWAPS are games or gamified platforms that serve a specific purpose which is not merely ludic. 

There is research on creating language resources using crowdsourcing, some of which are: 
Zombilingo for syntactic annotation \cite{fort2014creating}, Phrase detectives for co-reference annotation \cite{chamberlain2008phrase} or JeuxDeMots for the creation of a lexico-semantic network \cite{lafourcade2008jeuxdemots}.
However little work has been done on the combination of crowdsourcing and language \emph{learning}. 
Probably the most well-known approach on combining crowdsourcing and language learning was done by Duolingo \cite{garcia2013learning}, although besides the stated goal of ``translating the web while learning a language", it is not quite clear how the output is used. 
Recently, the use of implicit crowdsourcing techniques using language learners for the creation of language resources on par 
with expert-created content has also been explored \cite{nicolas2020creating}.

A related field of work is crowdsourcing for education, of which the closest sub-aspect pertaining to this work is the creation of educational content. Initiatives include for example crowdsourced textbook generation \cite{solemon2013review} or crowdsourcing video captioning correction by language learners to enhance learning \cite{culbertson2017have}. The interested reader is referred to \newcite{jiang2018review} for an extensive review of current literature and practices.



%% file: sections/3_data.tex
COCTAILL \cite{volodina2014you} is a corpus of coursebooks for Swedish as a second language that we used as the basis for identification of candidate multi-word expressions (MWEs) for this experiment. COCTAILL contains texts and exercises aimed at adult learners of Swedish, and covers five CEFR levels: A1, A2, B1, B2, C1, where A1 is beginner level and C1 is advanced level \cite{councilofeurope2001}, with several coursebooks at each level (see Table \ref{tab:data:coctaill}). In the corpus, each chapter (lesson) in a coursebook has been assigned a level at which it is known to be used in an L2 teaching context.
For example, suppose a textbook $T$ contains 9 chapters and that practicing teachers are using chapters $T1$-$T4$ when teaching students aiming for the A1 level, and chapters $T5$-$T9$ aiming for the A2 level. 
All texts that are used in chapters $T1$-$T4$ are surmised to target A1 level knowledge, while texts that are used in chapters $T5$-$T9$ are assumed to target A2 knowledge, and so on.
Further, all words that are used in the texts in chapters $T1$-$T4$ are labeled as potential target receptive vocabulary for the A1 level. All new vocabulary items that are used in texts in chapters $T5$-$T9$ (and that have not been used at previous levels) are labeled as potential target vocabulary at the A2 level, and so on. This approach allows us to generate useful vocabulary lists for both pedagogical and assessment use, as well as for automatic classifications of various kinds. However, generalizations about the levels at which vocabulary items should be targeted remains only an assumption that needs to be confirmed. Thus, the projected levels at the word level can serve as indications that certain items might be easier or harder, although we make no claims about the correctness of these projections. 

Table \ref{tab:data:coctaill} shows an overview of the corpus, detailing how many books targeting each CEFR level that are included, how many authors we rely on, 
 as well as the number of chapters, texts, sentences and tokens.

\begin{table*}[htbp]
    \centering
    \begin{tabularx}{\textwidth}{Xrrrrrr}
\toprule
CEFR level & \#Textbooks & \#Authors & \#Chapters & \#Texts & \#Sentences & \#Tokens \\ 
\midrule
A1 & 4 & 10 & 37 & 101 & 1581 & 11132 \\
A2 & 4 & 10 & 105 & 232 & 4217 & 37259 \\
B1 & 4 & 12 & 83 & 345 & 6510 & 79402 \\
B2 & 4 & 8 & 31 & 314 & 8527 & 101583 \\
C1 & 2 & 2 & 22 & 115 & 5085 & 71991 \\
Total & 18 & 42$^{*}$ & 278 & 1106 & 25920 & 301367 \\
\midrule
\multicolumn{7}{l}{\footnotesize $^{*}$ 26 unique} \\
\bottomrule
\end{tabularx}
    \caption{Statistics over COCTAILL per level}
    \label{tab:data:coctaill}
\end{table*}



COCTAILL is annotated automatically with the Sparv-pipeline\footnote{\neveranonurl{https://spraakbanken.gu.se/sparv}} \cite{borin2016sparv} for base forms, word classes, syntactic relations, word senses, MWEs and some other linguistic aspects. MWEs are identified on the basis of Saldo lexicon \cite{borin2013saldo} entries, which means that only MWEs that are contained in Saldo will be recognized. As Saldo is under active development, the automatic pipeline will probably be able to identify more MWEs in the future. 
From the annotated version of COCTAILL, we have generated a new version of the SVALex list \cite{franccois2016svalex} based on senses, 
as Sparv has been updated to include a word sense disambiguation module since the creation of the original list. Word sense distinctions are based on Saldo senses.

An entry in 
the list
consists of a combination of a base form with its word class (i.e. a lemgram), plus a word sense. Polysemous items have several distinct entries 
in the list and different frequency counts are associated with each of the sense entries. 
Each item 
contains its frequency distribution across different CEFR levels where it occurred and is associated with the lowest CEFR level of the texts in which it is observed. 
Starting from the list of 1351 MWEs in the 
list, two annotators classified them manually according to a custom typology \anoncitep{lindstromtiedemannINPREP}. 

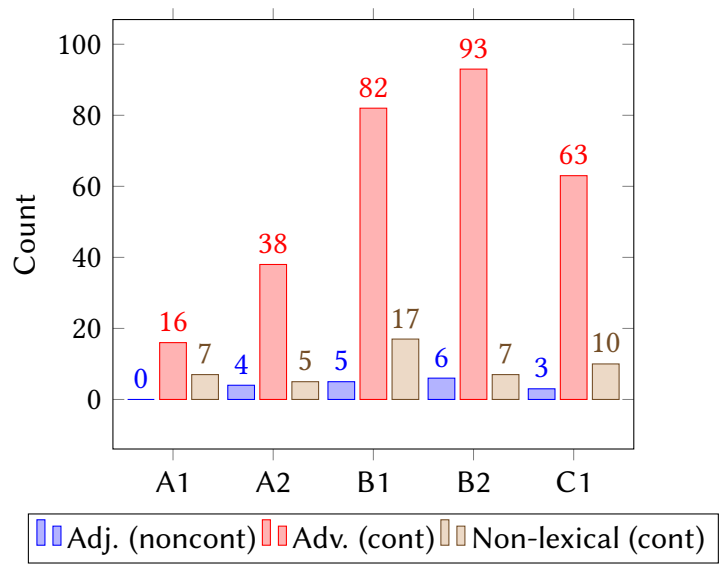
\begin{figure*}
\centering
\begin{tikzpicture}
\begin{axis}[
    ybar,
    enlargelimits=0.15,
    legend style={at={(0.5,-0.15)},
      anchor=north,legend columns=-1},
    ylabel={Count},
    symbolic x coords={A1,A2,B1,B2,C1},
    xtick=data,
    nodes near coords,
    nodes near coords align={vertical},
    ]
\addplot coordinates {(A1,0) (A2,4) (B1,5) (B2,6) (C1,3)};
\addplot coordinates {(A1,16) (A2,38) (B1,82) (B2,93) (C1,63)};
\addplot coordinates {(A1,7) (A2,5) (B1,17) (B2,7) (C1,10)};
\legend{Adj. (noncont), Adv. (cont), Non-lexical (cont)}
\end{axis}
\end{tikzpicture}
\caption{Group 3 in the crowdsourcing experiment}
\label{fig:data:group3}
\end{figure*}

For the experiment, we chose three different groups of MWEs based on this manual annotation, aiming to select a wide yet balanced variety of different types of expressions. 
This resulted in the selection of the following three groups: (1) interjections, fixed expressions and idioms,\footnote{We are aware of the difficulty of such distinctions. We tried to give strict definitions of fixed expressions and idioms as well as providing illustrative examples of both. However, comparisons of the annotations of the two annotators have shown that what annotator 1 classified as one of the categories could sometimes be annotated as one of the other categories by the other annotator which is why we decided to have these as a joint group for the experiment.} 
(2) verbal MWEs and (3) adverbial, adjectival and non-lexical MWEs. For the sake of conciseness and spatial limitations, we will refer to group 1 as ``interjections", to group 2 as ``verbs" and to group 3 as ``adverbs".
Figures \ref{fig:data:group1}, \ref{fig:data:group2} and \ref{fig:data:group3} show the number of occurrences per group per level in the resource based on the first round of annotation.
From each of these three groups, we selected 60 expressions to be used in the experiment, with 12 items for each CEFR level, for a total of $3 * 60 = 180$ expressions.
Expressions were de-contextualized in the sense that we did not provide any example sentences illustrating the use and context of the expression. While this decision may hinder the decision making process, it ensures that decisions are solely based on the expressions themselves, as opposed to syntactic complexity or other features that might be judged in a sentence.


\begin{figure*}
\centering
\begin{tikzpicture}
\begin{axis}[
    ybar,
    enlargelimits=0.15,
    legend style={at={(0.5,-0.15)},
      anchor=north,legend columns=-1},
    ylabel={Count},
    symbolic x coords={A1,A2,B1,B2,C1},
    xtick=data,
    nodes near coords,
    nodes near coords align={vertical},
    ]
\addplot coordinates {(A1,11) (A2,30) (B1,42) (B2,33) (C1,26)};
\addplot coordinates {(A1,3) (A2,12) (B1,20) (B2,32) (C1,37)};
\addplot coordinates {(A1,2) (A2,3) (B1,0) (B2,4) (C1,3)};
\legend{Fixed expr. (cont),Idioms (cont \& noncont),Interj. (cont)}
\end{axis}
\end{tikzpicture}
\caption{Group 1 in the crowdsourcing experiment}
\label{fig:data:group1}
\end{figure*}


\begin{figure*}
\centering
\begin{tikzpicture}
\begin{axis}[
    ybar,
    enlargelimits=0.15,
    legend style={at={(0.5,-0.15)},
      anchor=north,legend columns=-1},
    ylabel={Count},
    symbolic x coords={A1,A2,B1,B2,C1},
    xtick=data,
    nodes near coords,
    nodes near coords align={vertical},
    ]
\addplot coordinates {(A1,12) (A2,46) (B1,103) (B2,103) (C1,67)};
\addplot coordinates {(A1,5) (A2,15) (B1,29) (B2,21) (C1,24)};
\addplot coordinates {(A1,8) (A2,17) (B1,42) (B2,45) (C1,22)};
\legend{Particle verbs,Reflexive verbs,Other verbal MWEs}
\end{axis}
\end{tikzpicture}
\caption{Group 2 in the crowdsourcing experiment}
\label{fig:data:group2}
\end{figure*}
 
   
Within each of the groups we prioritized items that had been classified and agreed upon by both annotators. 
We double-checked all items in the COCTAILL corpus to see that the \emph{sense} we had listed was the one used in the corpus at the automatically assigned CEFR level; this step was necessary, as the automatic annotation of the corpus might not always identify the correct sense of a word or expression.

To make the experiment a learning experience and to make sure the level of difficulty was annotated in relation to a particular sense, we added definitions to all items. As far as possible we picked definitions from  \emph{Svensk ordbok} (\neveranonurl{svenska.se}). When this was not possible, we used Saldo, Wiktionary, Lexin, or provided definitions of our own.



%% file: sections/4_methodology.tex


Instead of having volunteers annotate each MWE with a target CEFR level (a task that requires in-depth knowledge of the CEFR), and following previous results showing that relative comparative judgments are easier than assigning items to a category \cite{lesterhuis2017comparative}, we opted to use best-worst scaling \cite{louviere2015best} for the crowdsourcing task. The rationale 
is that language proficiency is a continuum rather than a set of discrete proficiency levels, although for practical reasons it is simplified to a set of discrete levels \cite[p.~34]{coe2018common} (cf Section \ref{section:introduction}). 
Thus, using a relative ranking method may be more fruitful than trying to classify expressions into discrete classes; in addition, operating on a continuous scale allows for more sophisticated statistical measures to be used \cite[p.~131]{ortega2012interlanguage}.
    Further, using best-worst scaling we get a maximum amount of information with a minimal amount of clicks from the crowdsourcers 
 \cite{chrzan2019best}. Finally, such a set up requires no knowledge of the CEFR, as participants rely on their intuition when judging expressions against each other.


In best-worst scaling, one is presented with a group of items to rank and asked to rank one of the items as the “best” or the “easiest” and one of the items as the “worst” or the “hardest”. If one presents four items to the annotator to be ranked, having them choose both the easiest and the hardest out of the four expressions, it will result in 5 out of 6 possible relations.

To illustrate this further let us consider an example to show that four expressions give us six relations. Indeed, among four expressions, there exist six possible relations. Let us consider an example with expressions A, B, C and D, and let us assume that we want to know which of the expressions A, B, C and D that is the easiest and which is the hardest. This means that we thus have the following combinations of items between these expressions:

\begin{itemize}
\item $A B$
\item $A C$
\item $A D$
\item $B C$
\item $B D$
\item $C D$
\end{itemize}

As the relations are symmetrical, we do not need to consider other combinations such as B A, as it is identical to A B; saying that B is easier than A implies that A is harder than B. With best-worst scaling, if one chooses B as the easiest expression and C as the hardest expression, we have knowledge of the following relations:
\begin{itemize}
\item $B < C$
\item $B < A$
\item $B < D$
\item $C > A$
\item $C > D$
\end{itemize}
The first point is self-explanatory: as we have stated that B is easiest and C is hardest, B must be easier than C. The other relations follow logically. As we have declared B as the easiest item, B must be easier than any of the other items (points 2 and 3). As we have declared C to be the hardest item, it must be harder than all other items (points 4 and 5). The only relation that we do not have information about is the relation between items A and D. However, this relation will be covered by subsequent tasks in which 
A and/or D occur.

In order to cover all possible combinations using best-worst scaling, we have chosen a redundancy-reducing combinatorial algorithm to calculate the minimum amount of combinations of four items needed to cover all relations in such a way as to minimize redundancy, i.e. repeating items that have already been encountered, based on \anoncitet{cibej}.

With four items per task and 60 expressions there are 1,770 possible relations and 487,635 possible combinations.
Using the redundancy-reducing combinatorial algorithm, this means that we need to have 326 tasks.
Of the 1770 relations, 
\begin{itemize}
\item 1362 (77\%) are non-repetitive
\item 33 with 1 relation known
\item 50 with 2 relations known
\item 12 with 3 relations known
\item 3 with 4 relations known
\item 1 with 5 relations known
\end{itemize}
Thus  77\% of the relations are covered by non-repetitive combinations, while 23\% of the relations are covered by partially repetitive combinations.

Finally, using best-worst scaling leads to a decrease in effort spent on the task.
If one were to rank four items out of four in relation to each other, one would need at least four clicks, while best-worst scaling requires (a minimum of) two clicks, reducing the workload by half.

%% file: sections/5_0_experimental_setup.tex
One of the aims of this study is to test how one's background influences the outcome of a crowdsourcing experiment.
To take a step towards that aim, we experiment with two different ways of 
ranking MWEs according to difficulty.

\begin{enumerate}
\item Intuition-based (implicit) labeling, i.e. crowdsourcing: We ask a heterogeneous group of L2 speakers of Swedish (non-experts) as well as experts (L2 Swedish professionals e.g. teachers, researchers) to rank items by taking part in a crowdsourcing experiment where we subdivide the expert group into a general L2 professional group and a group of CEFR-experts:
    \begin{itemize}
        \item Non-experts: L2 speakers of Swedish 
        at intermediate level (B1) or above (according to self-assessment)
        \item Experts -- L2 Professionals: Teachers, assessors and/or researchers of Swedish as a second language (referred to as L2 professionals)
        \item Experts -- CEFR experts: A separate subgroup of L2 professionals who use CEFR in their L2 Swedish practices
    \end{itemize}
\item Expert judgment-based (explicit) labeling: We ask a small group of CEFR experts (teachers/researchers/assessors) to label MWE items manually for the levels at which they expect L2 learners to understand them. This annotation task is formulated in \textit{levels} rather than \textit{relative ordering} to resemble a real-life annotation scenario as much as possible where experts would be involved -- which, however, entails some difficulties in comparison of the results.
 \end{enumerate}

\subsection{Practicalities}

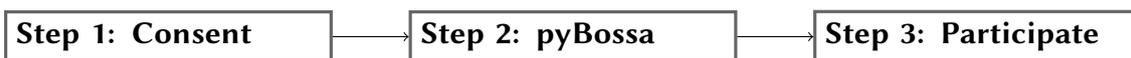
\begin{figure*}[htbp]
    \centering
    \begin{tikzpicture}[squarednode/.style={rectangle, draw=black!60, very thick, minimum size=5mm},]
    \node[squarednode,text width=4cm]      (startnode)                              {\textbf{Step 1: Consent}
    };
    \node[squarednode,text width=4cm]      (secondnode)       [right=of startnode] {\textbf{Step 2: pyBossa}
    };
    \node[squarednode,text width=4cm]      (rightsquare)       [right=of secondnode] {\textbf{Step 3: Participate}
    };
    
    \draw[->] (startnode.east) -- (secondnode.west);
    \draw[->] (secondnode.east) -- (rightsquare.west);
    \end{tikzpicture}
    \caption{Practical steps for participants in the crowdsourcing experiment}
    \label{fig:practical:steps}
\end{figure*}

Figure \ref{fig:practical:steps} illustrates the steps necessary to take part in the experiment.
In the first step of the experiment, to comply with the GDPR \cite{eu2016gdpr} we asked our participants for consent to use their background information for this research and to send out gift certificates.\footnote{Expert form (Swedish only):\\\url{https://spraakbanken.gu.se/larkalabb/mwe-cs-annotation-teacher}\\Non-expert form (Swedish only):\\\url{https://spraakbanken.gu.se/larkalabb/mwe-cs-annotation-crowd}}
At the same time, we collected information about the linguistic background as well as some other demographic variables as illustrated in Table \ref{tab:demographic}.

After filling out the consent form, participants were provided with guidelines and links for the crowdsourcing part of the experiment in the form of an automated email sent to the email address specified in the consent form.\footnote{Guidelines (in Swedish): \url{https://docs.google.com/document/d/1E700mnqaZ15cHr_3gXMvg0d4onm2ncMf36t-KUQuRrw/}} The guidelines were intentionally provided only in Swedish as a ``selection" principle to exclude L2 speakers of lower proficiency levels. 

In the second step, participants were asked to create an account on the crowdsourcing platform, with the explicit instruction to use the same email address as provided in the consent form so that we could link their background information 
to the crowdsourcing results. Email addresses were solely collected for this purpose and were discarded after this linking step was performed.

As a final step, participants were asked to participate in the projects proper.
Each crowdsourcer was expected to complete at least 84 
items out of 326 in each of the three projects, which amounted to a total time of about 30-45 minutes per project. Participants who completed at least 84 tasks per project were sent a gift (step 3 in Figure \ref{fig:practical:steps}).\footnote{In the later stages of the experiment when it was not possible to contribute 84 tasks in one or more projects, we relaxed the constraints for gift eligibility to $\approx$~240 tasks in total.} 

To reach the crowdsourcing population, we published announcements via e-mail, social networks, and through professional and private networks. For CEFR experts, we listed requirements with regards to their qualifications and recruited three experts on the basis of this.

We left a calendar month for the crowdsourcing experiment from the date of the first announcement, with periodic reminders to recruit broader participation. All crowdsourcers that met our requirements of the minimal contributions, were sent small gifts. Experts were paid by the hour.

\subsection{Implementation}
For the crowdsourcing experiment, we set up nine projects for the three different participant backgrounds. All projects were implemented in pyBossa, an open-source customizable framework for crowdsourcing tasks developed by SciFabric.\footnote{\neveranonurl{https://pybossa.com/}} For each of our three target groups (Non-experts = L2 speakers, L2 Professionals = L2 teachers, researchers; CEFR Experts = L2 teachers, researchers, assessors with CEFR experience) we prepared three projects consisting of three sets of different MWE-types (3 participant groups x 3 projects = 9 crowdsourcing projects). In addition, we set up a tenth crowdsourcing experiment for people who did not conform to any of the three target groups or for people who wanted to see how the projects work.\footnote{Test-project (in Swedish): \url{https://ws.spraakbanken.gu.se/ws/tools/crowd-tasking/project/l2p_mwe_group2_other/}}

\begin{figure*}
    \centering
    \includegraphics[width=.8\textwidth]{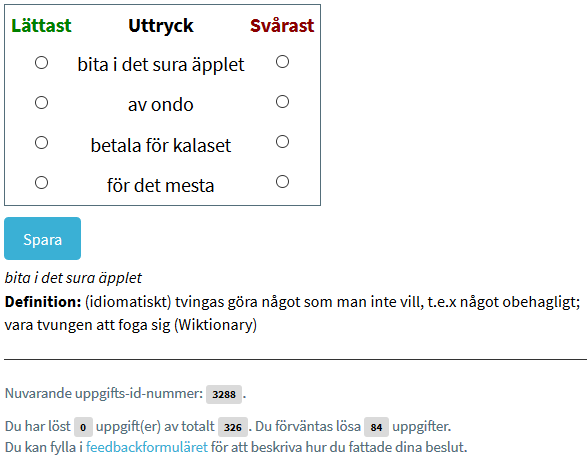}
    \caption{Example of an MWE ranking task in pyBossa (lättast = easiest, svårast = most difficult, uttryck = expression; spara = save)}
    \label{fig:pybossa:gui}
\end{figure*}

For each of the projects, we arranged 
the 60 selected items per MWE group in such a way that the crowd could vote on their relative difficulty.
Figure \ref{fig:pybossa:gui} shows the graphical user interface we designed for this task, based on \anoncitet{cibej}. 
In the user interface, crowdsourcers were shown four MWEs and were asked to indicate which expression they found the easiest and the hardest to understand by using the buttons on the left and the right of the expressions, based on their own intuition. In addition, one could click on any of the four expressions to be shown a definition in case one 
was not sure about 
the meaning of an expression.
The interface also showed a pyBossa-internal ID number, the number of tasks that had been completed by the crowdsourcer, the number of total tasks (326 for each project) and the expected number of tasks that each crowdsourcer should finish (84 for each project, except for the ``CEFR experts" who were expected to complete all 326 tasks). 
Finally, we also included a link to a feedback form where crowdsourcers could indicate their reasoning about assigning the labels for easiest and hardest, or any other feedback they may wish to provide. 

As additional safe-guards, we implemented checks for user errors for the following cases:

\begin{enumerate}\setlength\itemsep{0em}
    \item No value selected
    \item Only one column is selected
    \item Same value in both columns
\end{enumerate}
As we wanted to collect the easiest and the hardest expression among a set of four expressions, it was disallowed not to provide any value (point 1), to only choose either an easiest expression or a hardest expression but not both (point 2) or to select the same expression as both the easiest and the hardest (point 3).
Furthermore, as we wanted to maximize user interaction,
we took care to make sure that the platform was functional and usable not only on desktop PCs but also on smaller screens such as smartphones. By doing so, people could use their smartphones wherever they were and whenever they had a minute to continue working on the tasks.  
Other considerations concerned the placement of the ``easiest" and ``hardest" columns, color schemes, and the ease of use on a smart phone.
After registration in pyBossa participants could log in and continue from where they left off at any time suitable to them and on any platform (smartphone, tablet, computer).
As to the number of votes per task, i.e. how many different answers were needed per task for a task to be considered complete, we set the number to 5 for L2 speakers and to 3 for L2 professionals and CEFR experts. These numbers were picked based on the estimated number of participants in the various groups. This meant that each single task in the project would have 5 respectively 3 answers (i.e. judgments about the easiest and the hardest expression) by 
different annotators.

We assigned the following scores to expressions: 1 for the expression that was rated as the easiest, 3 for the expression that was rated as the hardest and 2 for the two unrated expressions.

\subsection{Experimental design}

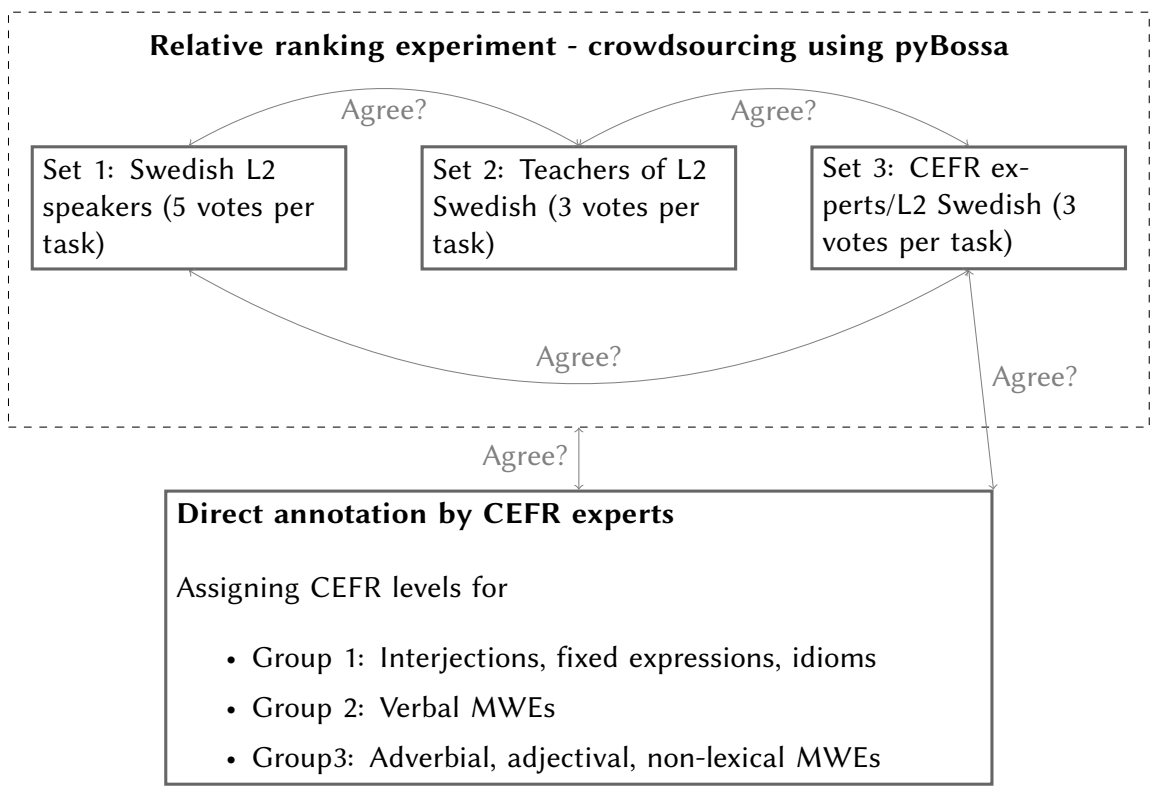
\begin{figure*}
    \centering
    \resizebox{.95\textwidth}{!}{
    \begin{tikzpicture}[squarednode/.style={rectangle, draw=black!60, very thick, minimum size=5mm,text width=4cm,}]
    \node[squarednode] (in1) {Set 1: Swedish L2 speakers (5 votes per task)};
    \node[squarednode] (in2) [right=of in1] {Set 2: Teachers of L2 Swedish (3 votes per task)};
    \draw[bend left,<->,gray]  (in1.north) to node [below] {Agree?} (in2.north);
    \node[squarednode] (in3) [right=of in2] {Set 3: CEFR experts/L2 Swedish (3 votes per task)};
    \draw[bend left,<->,gray]  (in2.north) to node [below] {Agree?} (in3.north);
    \draw[bend right,<->,gray]  (in1.south) to node [above] {Agree?} (in3.south);
    \node[text width=15cm,align=center] (tnode1) [above=of in2] {\textbf{Relative ranking experiment - crowdsourcing using pyBossa}};
    \node[minimum height=1cm] (dummy1) [below=of in2] {};
    \node[draw,dashed,fit=(in1) (in2) (in3) (tnode1) (dummy1)] (out1) {};
    
    \node[rectangle, draw=black!60, very thick, minimum size=5mm,text width=11cm] (in4) [below=of dummy1] {\textbf{Direct annotation by CEFR experts}\\~\\
    Assigning CEFR levels for\\\begin{itemize}
    \setlength\itemsep{0em}
        \item Group 1: Interjections, fixed expressions, idioms
        \item Group 2: Verbal MWEs
        \item Group3: Adverbial, adjectival, non-lexical MWEs
    \end{itemize}};
    
    \draw[<->,gray] (in4.north east) to node [right] {Agree?} (in3.south);
    \draw[<->,gray] (in4.north) to node [auto] {Agree?} (out1.south);
    \end{tikzpicture}%
    }
    \caption{Overview of the experimental design}
    \label{fig:agreements}
\end{figure*}

Figure \ref{fig:agreements} shows an overview of the experimental design. In the experiment, we wanted so see whether
non-experts and experts agree with each other about the relative difficulty of multiword expressions in the crowdsourcing experiment. We further subdivided the expert group into two to be able to compare experts' indirect judgement (crowdsourcing) to their direct (explicit) labeling, but this was only done with the small subgroup of CEFR experts who we therefore had to make sure were all well familiar with CEFR in connection with their work.
Finally, we wanted to check whether individual explicit labels by the CEFR experts coincided with the group results from their implicit crowdsourcing experiment.

As indicated above we also asked our three CEFR experts to perform a direct labeling task. 
This meant that we asked them to go through all of the selected MWEs in a spreadsheet and decide at which CEFR level these MWEs could be expected to be understood. All three CEFR experts were asked to do the crowdsourcing experiment in pyBossa first and were only given access to the spreadsheet for direct labeling after they had completed that, to make their crowdsourcing experience as similar to that of the rest as possible. However, unlike the other participants the CEFR experts were asked to rank all items in all the three pyBossa projects ($3*326$). In the direct annotation experiment, they were asked to pick one level from a drop-down menu with A1, A2, B1, B2, C1, C2 or above for each item in a spreadsheet with all 180 MWEs.

\subsection{Demographic information}
\input{sections/5_4_demographic.tex}

\subsection{Evaluation methodology}
\input{sections/5_5_evaluation.tex}

%% file: sections/5_4_demographic.tex
To better understand whether and/or how the intuitions and judgments were influenced by the background of the participants, we collected information about our participants in a separate form (personal metadata). Since L2 Swedish is widely spread in Sweden and Finland, these two countries were our primary targets. However, we used social media and our personal professional networks to spread information about the experiment, which also encouraged participation from other countries. Out of 79 consent registrations in total, 50 crowdsourcers participated in the experiment, which constitutes a  drop-out rate of 37\%.
Upon completing the crowdsourcing experiment, we could see the following participant characteristics (Table 2):

We attracted 27 L2 non-experts (L2 speakers) and 23 L2 experts, 
including the three CEFR experts. 
Sweden and Finland contributed with 22 participants each (at 44\%). The first language of the contributors is dominated by Finnish (30\%), but other first languages are also represented, including Swedish (20\%), German (12\%), Russian (8\%), Spanish (4\%), Arabic (4\%), Hungarian (4\%) and others. The population is well-educated having either a pre-doctoral university degree (60\%)  
or a doctoral degree (36\%). L2 speakers provided self-assessed levels of Swedish as B1 or above in 96\% of the cases, with one outlier at the A1 level. 65\% of the L2 experts 
have 10 years or more of experience of teaching or assessing Swedish as a second language.  The age characteristics show that we attracted a rather ``mature" population (78\%), whereas people of 30 years and younger are less represented (22\%). 
The gender representation is rather unbalanced 
(66\% women versus 28\% men with 6\% who preferred not to answer that question), which can be due to a recruitment bias or -- potentially -- reflect gender representation within the areas of language learning and teaching.  All in all, we have participants of various background profiles, which represents the target group for the intended output of the research.

The three CEFR experts recruited come from Finland since Finland appears to use CEFR more extensively in the teaching and assessment of L2 Swedish than Sweden does. All CEFR experts have Finnish as their L1 and represent: one L2 Swedish teacher, one L2 Swedish researcher (PhD) and one L2 Swedish assessor (PhD).

\clearpage
\onecolumn
\begin{landscape}
\begin{longtable}{llp{44pt}p{45pt}p{45pt}p{45pt}p{45pt}p{45pt}p{50pt}p{50pt}}
\toprule
\caption{Demographic variables} \\
 & Profiles & L2 speakers & L2 experts & Finland L2 speakers & Finland L2 experts & Sweden L2 speakers & Sweden L2 experts & Other L2 speakers & Other L2 experts \\ \midrule \endfirsthead
\caption{Demographic variables continued} \\
& Profiles & L2 speakers & L2 experts & Finland L2 speakers & Finland L2 experts & Sweden L2 speakers & Sweden L2 experts & Other L2 speakers & Other L2 experts \\ \midrule \endhead
Total & 50 & 27 & 23 & 9 & 13 & 13 & 9 & 5 & 1 \\ \midrule
\textbf{Gender} & & & & & & & & & \\
Female & 33 & 15 & 18 & 7 & 10 & 7 & 8 & 1 & - \\
Male & 14 & 9 & 5 & 2 & 3 & 4 & 1 & 3 & 1 \\
Other & 3 & 3 & - & - & - & 2 & - & 1 & - \\ \midrule
\textbf{Age} & & & & & & & & & \\
16-20 & 5 & 5 & - & 3 & - & 2 & - & - & - \\
21-30 & 6 & 2 & 4 & 2 & 4 & - & - & - & - \\
31-40 & 15 & 10 & 5 & 4 & 3 & 3 & 2 & 3 & - \\
41+ & 24 & 10 & 14 & - & 6 & 8 & 7 & 2 & 1 \\ \midrule
\textbf{Education} & & & & & & & & & \\ 
High school & 2 & 2 & - & - & - & 2 & - & - & - \\
University & 30 & 16 & 14 & 9 & 8 & 5 & 6 & 2 & - \\
PhD & 18 & 9 & 9 & - & 5 & 6 & 3 & 3 & 1 \\ \midrule
\multicolumn{2}{l}{\textbf{Mother tongue}} & & & & & & & & \\ 
Arabic & 2 & 2 & - & - & - & 2 & - & - & - \\
Dutch & 1 & 1 & - & - & - & - & - & 1 & - \\
English & 1 & 1 & - & - & - & 1 & - & - & - \\
Finnish & 17 & 8 & 9 & 8 & 9 & - & - & - & - \\
German & 6 & 5 & 1 & 1 & - & 3 & - & 1 & 1 \\
Hungarian & 2 & 2 & - & - & - & 1 & - & 1 & - \\
Luxembourgish & 1 & 1 & - & - & - & 1 & - & - & - \\
Norwegian & 1 & 1 & - & - & - & 1 & - & - & - \\
Russian & 4 & 3 & 1 & - & - & 3 & 1 & - & - \\
Serbian & 2 & - & 2 & - & 1 & - & 1 & - & - \\
Slovenian & 1 & 1 & - & - & - & - & - & 1 & - \\
Spanish & 2 & 2 & - & - & - & 1 & - & 1 & - \\
Swedish & 10 & - & 10 & - & 3 & - & 7 & - & - \\ \midrule
\textbf{L2 Swedish level} & & & & & & & & & \\ 
A1 & 1 & 1 & - & - & - & - & - & 1 & - \\
A2 & - & - & - & - & - & - & - & - & - \\
B1 & 4 & 4 & - & 2 & - & - & - & 2 & - \\
B2 & 7 & 7 & - & 2 & - & 4 & - & 1 & - \\
C1 & 9 & 9 & - & 4 & - & 4 & - & 1 & - \\
C2 & 6 & 6 & - & 1 & - & 5 & - & - & - \\ \midrule
\textbf{Teaching years} & & & & & & & & & \\ 
1-9 & 8 & - & 8 & - & 6 & - & 2 & - & - \\
10-19 & 7 & - & 7 & - & 3 & - & 3 & - & 1 \\
20+ & 4 & - & 4 & - & 1 & - & 3 & - & - \\
other & 4 & - & 4 & - & 4 & - & 1 & - & - \\
\bottomrule
\end{longtable}\label{tab:demographic}
\end{landscape}
\clearpage
\twocolumn

%% file: sections/5_5_evaluation.tex
We use two modes of annotating items for their difficulty: crowdsourcing by non-experts and experts, and direct annotation by experts. 
Since direct expert annotation is a rather traditional approach, we use traditional ways of evaluating it, relying on metrics such as agreement and Spearman rank correlation. However, crowdsourcing is a new approach for this type of tasks, thus we explain how we evaluate and compare the results of the crowdsourcing experiment below. The results are  presented in Section 6. 


For evaluation of the crowdsourcing, we project each expression onto a linear scale.
The scale ranges from 1 to 3, with a minimum value of 1 if an expression was always classified as the easiest out of a set of four possible expressions and a maximum value of 3 if it was always classified as the most difficult out of a set of four possible expressions by all annotators. Otherwise, the expression \emph{exp} is assigned the score $s(exp)$ as the mean of all assigned scores $x$ according to the formula
\begin{equation}
s(exp) = \frac{\sum_{i=1}^{n}x_i}{n}    
\end{equation}

With $x_i$ being the $i$-th score assigned to $exp$ and $n$ being the total number of scores assigned to $exp$. 
The limits of 1 and 3 are predetermined by our way of measuring “the easiest” and “the hardest” expression with best-worst scaling (cf Section \ref{section:methodology}). 

We sort the data according to the reverse order of $s(exp)$ and assign sequential ranks from 1 to 60 to the resulting ordering.


%% file: sections/6_0_results.tex
In this section we present the results from the different experiments.
First, we look into the results from the crowdsourcing experiment. 
After this we look into the results from the direct annotation (expert labeling) which we also compare to the rankings which this group of experts did in pyBossa. 
Finally, we also investigate how the number of votes influences the results and how much time is needed for the crowdsourcing experiment as opposed to direct expert annotations.

\subsection{Linear scale}
The experiments generated rich data for analysis. In this section we look at the results of the study from a quantitative point of view. For the purpose of comparison, we projected results of crowd-votings to linear scales based on the fact that each vote in a crowdsourcing task assigns scores to the items: either 1 (``easiest"), 3 (``most difficult"), or 2 for each of the two items in-between. Based on the numerical values, all items are listed in the order of their scores corresponding to the perceived degree of difficulty. 

Based on that principle, we obtained one linear scale per participant group and one representing the whole population of crowdsourcers (mixed background rankings). 

Table \ref{tab:results:linear:scale:spearman} shows the Spearman rank correlation coefficient between the three sets of MWEs and the three groups of participants. Spearman rank correlation coefficient has a range from -1 to +1 where -1 indicates a perfect negative correlation; zero indicates no correlation; and +1 indicates perfect positive correlation.

\begin{table*}[htbp]
    \centering
    \begin{tabular}{lccc}
        \toprule
         & Gr.1 (interj.) & Gr.2 (verbs) & Gr.3 (adv.)\\
         \midrule
         L2 speakers-L2 professionals & 0.9509 & 0.9282 & 0.9203 \\
         L2 speakers-CEFR experts & 0.9333 & 0.8115 & 0.8370 \\
         L2 professionals-CEFR experts & 0.9386 & 0.8495 & 0.8579 \\
         \bottomrule
    \end{tabular}
    \caption{Agreement between voter groups in the crowdsourcing experiment}
    \label{tab:results:linear:scale:spearman}
\end{table*}

As can be gathered from Table \ref{tab:results:linear:scale:spearman}, the highest correlations can be found between non-experts (here meaning L2 speakers/learners) 
and the general group of ``L2 professionals'' (including teachers, assessors, researchers) across all of the three MWE groups, while the correlations between non-experts (L2 speakers) and ``CEFR-experts'' (i.e. the subgroup of three L2 professionals) are the lowest among all the three MWE groups. We can thus say that non-experts (L2 speakers) and experts (L2 professionals) in our experiment agree very well on the relative difficulty of MWEs, followed by L2 professionals and CEFR experts, while L2 speakers and CEFR experts tend to agree to a lesser extent. 
Despite these marginal fluctuations, we can see strong correlations between all of the tested target groups across all the three sets of tested MWEs. This indicates that intuitions about the difficulty of MWEs are more or less shared across all tested groups, despite the differences in background and professional competence. It seems that we can confirm that non-experts -- that is, L2 speakers lacking expertise and competence in a subject (e.g. language assessment) -- can be seen as on par with experts for tasks requiring high competence, something that has also been shown in approaches in citizen science \cite{kullenberg2016citizen}. 

To get an insight into how well individuals can agree on crowdsourcing tasks we looked at the three CEFR experts in our experiment who completed the full sets of tasks in all of the three pyBossa projects. Table \ref{tab:results:linear:scale:experts:iaa} shows the Spearman rank correlation based on their individual linear scales calculated from the crowdsourced data. As can be seen from Table \ref{tab:results:linear:scale:experts:iaa}, annotators 1 and 2 tend to agree the most, while annotators 1 and 3 tend to agree the least, with annotators 2 and 3 falling in-between. This might be a result of their different backgrounds and how often they use CEFR explicitly. The more voters we have, the less bias there is in the resulting data \cite[e.g.][]{snow2008cheap}.

\begin{table*}[htbp]
    \centering
    \begin{tabular}{cccc}
        \toprule
         & Gr.1 (interj.) & Gr.2 (verbs) & Gr.3 (adv.)\\
         \midrule
         CEFR experts 1 and 2 & 0.8130 & 0.8581 & 0.7735 \\
         CEFR experts 1 and 3 & 0.7733 & 0.5788 & 0.6988 \\
         CEFR experts 2 and 3 & 0.7964 & 0.6236 & 0.7026 \\
         \bottomrule
    \end{tabular}
    \caption{Inter-annotator agreement for CEFR experts in the crowdsourcing experiment calculated with Spearman rank correlation coefficient}
    \label{tab:results:linear:scale:experts:iaa}
\end{table*}

\subsection{Expert labeling}
If we look closer at the simple and extended percentage agreement between the CEFR expert annotators in the explicit (interchangeably called `direct') labeling experiment, we can see that agreement is generally quite low for simple agreement (Tolerance 0 in Table \ref{tab:results:expert:direct:agreement}). With a tolerance of zero, one counts exact agreement between the annotators (e.g. the same item has been assigned to the same CEFR level). However, if one relaxes the tolerance level to 1 (extended percentage agreement), meaning that positive agreement also includes cases where annotators differed by only one level (e.g. one annotator said the item was A2 while another annotator said the item was B1), we can see that agreement drastically improves, as illustrated in Table \ref{tab:results:expert:direct:agreement}.

\begin{table*}[htbp]
    \centering
    \begin{tabular}{llll}
    \toprule
         &  Group 1 (interjections) & Group 2 (verbs) & Group 3 (adverbs) \\
         \midrule
         Tolerance 0 & 15.00 & 21.70 & 13.30 \\
         Tolerance 1 & 61.70 & 58.30 & 65.00 \\
         \bottomrule
    \end{tabular}
    \caption{Agreement between CEFR experts in a direct labeling experiment in percent}
    \label{tab:results:expert:direct:agreement}
\end{table*}

In general, this gives us a picture that expert judgments are not ideal and that reaching an exact agreement between them is possibly an unattainable target, which also confirms the results from essay evaluation according to the CEFR-scale as presented in e.g. \newcite{diez2012use}. Given that direct labeling is a subjective and cognitively challenging task, more opinions than one are required \cite[cf][]{snow2008cheap,carlsen2012proficiency}. 
The MWEs in the experiments are de-contextualized which might further complicate decisions. 
This speaks in favor of assuming tolerance level 1 since the assigned levels describe a continuum of proficiency rather than strict categories \cite[p.~34]{coe2018common}. A hypothesis in connection to this is that disagreement outside tolerance 1 may indicate items that are on the periphery of the lower CEFR level, while items within tolerance 1 constitute the core vocabulary on the lower level. This is something to be explored in future research. 

Results of agreement between the explicit ranking of each individual expert and their own individual implicit judgment from the crowdsourcing experiment based on a comparison of the linear scales show mixed results (Table \ref{tab:results:expert:self}).  

\begin{table*}[htbp]
    \centering
    \begin{tabular}{llll}
    \toprule
         & Group 1 (interjections) & Group 2 (verbs) & Group 3 (adverbs) \\ 
         \midrule
         Expert 1 & 0.9095 & 0.9280 & 0.8935 \\
         Expert 2 & 0.8483 & 0.6147 & 0.7299 \\
         Expert 3 & 0.8010 & 0.5248 & 0.5540 \\
         \bottomrule
    \end{tabular}
    \caption{Spearman rank correlation coefficients for intra-annotator agreement between implicit and explicit modes of annotation}
    \label{tab:results:expert:self}
\end{table*}

Expert 1 is very consistent in both annotation methods, and all annotators seem to agree with themselves most for MWE group 1, while other agreements are lower. This could indicate that expert 1 is the one with the most experience with working with CEFR-levels. 
The inconsistency of the results for the same expert indicates that the expert reasons differently when using different methods, and that the way of reasoning influences the results. It has been previously shown that explicit scoring is more subjective and cognitively demanding than assessing by comparing two samples to each other \cite{lesterhuis2017comparative}, which also seems to be confirmed in this experiment. 
This indicates that we should not compare the two types of annotation and that expert judgment can only give reliable annotation if a reasonably large number of experts is used to counter-balance a potential subjective bias (cf. \newcite{snow2008cheap}). 
How large a number constitutes a ``reasonable amount'' is still an open question.

\subsection{Number of votes}
\input{sections/6_5_number_votes.tex}
\label{section:number:votes}

\subsection{Time investment}
\input{sections/6_6_time_invest.tex}

\label{section:time:investment}

%% file: sections/6_5_number_votes.tex
\newcite{aker2012assessing}  found that using one set of non-expert results (results from different annotators) outperformed using one single non-expert’s results, as the diversity of the crowd might cancel a high bias present in a single annotator.
In order to see how the number of votes influences the results, we randomly selected votes for the sample sizes 1, 2 and 3 (for the non-expert crowd, for which we collected 5 votes) and derived the linear scales, for each group separately as well as a randomly sampled mixed version over all three groups (`Mixed' in Tables \ref{tab:results:linear:scale:oop} and \ref{tab:results:linear:scale:d}). We then compared the linear scales of the different sample sizes to the linear scale derived from the full set votes (3 for experts and 5 for non-experts; for the mixed group we calculated the target linear scale from a random sample of three votes from both experts and non-experts), meaning that we compare for example the linear scale for non-experts derived from a single vote versus the linear scale for non-experts derived from 5 votes; the linear scale for non-experts derived from two votes versus the linear scale for non-experts derived from 5 votes; or the linear scale derived from randomly sampling two votes from both experts and non-experts versus the linear scale derived from randomly sampling three votes over all groups. 

\begin{figure*}[htbp]
    \centering
        \begin{tikzpicture}
    \node (Z) at (5,5) {$l_1$};
    \node (Y) at (10,5) {$l_2$};
    
    \node (X) at (2,1) {Least frequent};
    \node (W) at (2,4) {Most frequent};
    
    \node (S) at (13,5) {Rank difference};
    
    \node (G) at (5,4) {$A$};
    \node (H) at (10,4) {$D$};
    \node at (13,4) {1};
    
    \node (A) at (5,3) {$B$};
    \node (B) at (10,3) {$A$};
    \node at (13,3) {2};
    
    \node (C) at (5,2) {$C$};
    \node (D) at (10,2) {$C$};
    \node at (13,2) {0};
    
    \node (E) at (5,1) {$D$};
    \node (F) at (10,1) {$B$};
    \node at (13,1) {3};
    
    \draw (G) -- (B);
    \draw (A) -- (F);
    \draw (C) -- (D);
    \draw (E) -- (H);
    
    \end{tikzpicture}
    \caption{Out-of-place metric illustration}
    \label{fig:oop:illustration}
\end{figure*}
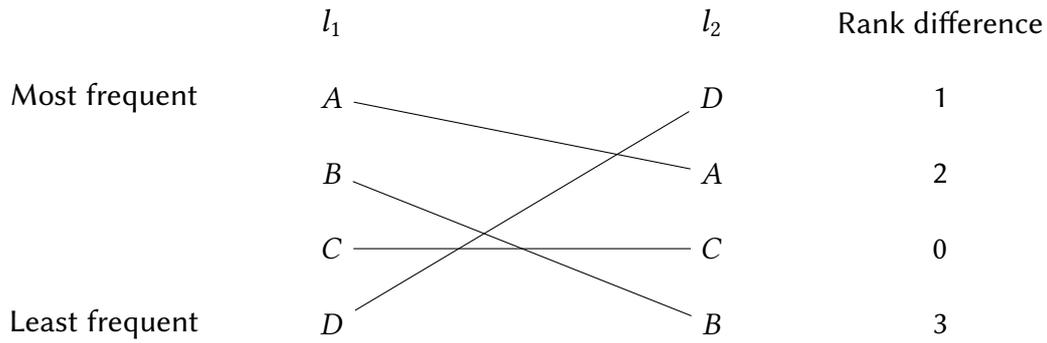

In order to quantify the differences between the scales, we used the \emph{out-of-place metric} $m_{oop}$ \cite{cavnar1994n}. This is a straightforward metric that measures the difference between two ranked lists and quantifies the difference. The reason for choosing this metric over rank correlation measures is that Spearman's correlation coefficient was very high and had similar values across all comparisons (see Table \ref{tab:results:linear:scale:oop}). While a high correlation is a positive result in itself, it does not allow for a detailed analysis. We surmise that using $m_{oop}$ may give a more tangible result. 
It is formalized as shown in (\ref{eq:oop})

\begin{equation}
    m_{oop} = \sum_{i=1}^n(|r(x_i,l_1) - r(x_i,l_2)|)
    \label{eq:oop}
\end{equation}

\noindent with 
$n$ being the number of items in the lists (the lists to be compared are of the same length in our case), $x_i$ being the $i$-th item, $r(x_i,l_1)$ being the rank of $x_i$ in the first list and $r(x_i,l_2)$ being the rank of $x_i$ in the second list.
To illustrate this, let us consider two lists $l_1$ and $l_2$ both containing the expression $A, B, C$ and $D$, but at different ranks. Figure \ref{fig:oop:illustration} shows a hypothetical scenario. In order to obtain $m_{oop}$, one first calculates the difference in ranks between the expressions, then sums up the differences. Thus, in this example, we would have $m_{oop} = 1 + 2 + 0 + 3 = 6$.
We also calculate how many items are at the exact same rank in both lists (out of 60 total).


\begin{table*}[htbp]
    \centering
    \begin{tabularx}{\textwidth}{XXXXXX}
    \toprule
         MWE group & Crowd & Sample size & $m_{oop}$ & $\rho$ & Same rank \\ 
         \midrule
         Interj. & L2 sp. & 1 & 150 & 0.98 & 8 \\
          &  & 2 & 112 & 0.98 & 15 \\
          &  & 3 & 102 & 0.99 & 16 \\
          & L2 prof. & 1 & 160 & 0.97 & 16 \\
          &  & 2 & 82 & 0.99 & 15 \\
          & CEFR exp. & 1 & 114 & 0.98 & 18 \\
          &  & 2 & 80 & 0.99 & 18 \\ 
          & Mixed & 1 & 114 & 0.98 & 14 \\
          & & 2 & 78 & 0.99 & 23 \\
         \midrule
         Verbs &  L2 sp. & 1 & 256 & 0.94 & 6 \\
          &  & 2 & 172 & 0.97 & 6 \\
          &  & 3 & 114 & 0.98 & 18 \\
          & L2 prof. & 1 & 196 & 0.97 & 8 \\
          &  & 2 & 90 & 0.99 & 18 \\
          & CEFR exp. & 1 & 200 & 0.96 & 7\\
          &  & 2 & 120 & 0.98 & 14 \\ 
          & Mixed & 1 & 138 & 0.98 & 11 \\
          & & 2 & 70 & 0.99 & 26 \\
         \midrule
         Adverbs &  L2 sp. & 1 & 254 & 0.95 & 4 \\
          &  & 2 & 154 & 0.98 & 12 \\
          &  & 3 & 110 & 0.98 & 15 \\
          &  L2 prof. & 1 & 244 & 0.94 & 8 \\
          &  & 2 & 132 & 0.98 & 14 \\
          & CEFR exp. & 1 & 126 & 0.98 & 14\\
          &  & 2 & 106 & 0.99 & 13\\ 
          & Mixed & 1 & 128 & 0.98 & 14 \\
          & & 2 & 54 & 0.99 & 25 \\
         \bottomrule
    \end{tabularx}
    \caption{Out-of-place calculations, Spearman's $\rho$ and same rank number for different numbers of votes}
    \label{tab:results:linear:scale:oop}
\end{table*}

We find that each of the sub-sampled lists compared to the full-vote list yields high Spearman rank correlation coefficients, with Spearman's $\rho$ varying from $\rho = 0.941, p = 5^{-29}$ to $\rho = 0.997, p = 3^{-66}$.
As can be gathered from Table \ref{tab:results:linear:scale:oop}, group 1 (interjections) 
shows the least amount of divergence among all three MWE groups, but also among the different crowds. 
Further, it can be observed that sampling over all three crowd groups produces more stable results than within-group sampling.

\begin{table*}[htbp]
    \centering
    \begin{tabularx}{\textwidth}{XlXXXXXXX}
    \toprule
        MWE group & Crowd & Sample size & \multicolumn{6}{c}{$d$} \\
        & & & 0 & 1 & 2 & 3 & 4 & 5 \\
        \midrule
         Interj. & L2 sp. & 1 & 8 & 21 & 34 & 45 & 49 & 55 \\
         & & 2 & 15 & 30 & 43 & 48 & 51 & 56 \\
         & & 3 & 16 & 34 & 44 & 51 & 56 & 58 \\
         & L2 prof. & 1 & 16 & 30 & 38 & 43 & 47 & 52 \\
         & & 2 & 15 & 38 & 53 & 56 & 58 & 59 \\
         & CEFR exp. & 1 & 18 & 35 & 42 & 46 & 53 & 57 \\
         & & 2 & 18 & 42 & 49 & 55 & 57 & 59 \\
         & Mixed & 1 & 14 & 30 & 44 & 49 & 55 & 57 \\
         & & 2 & 23 & 37 & 49 & 54 & 59 & 60 \\ 
         \midrule
         Verbs & L2 sp. & 1 & 6 & 10 & 27 & 36 & 42 & 45 \\
         & & 2 & 6 & 26 & 37 & 42 & 48 & 52 \\
         & & 3 & 18 & 34 & 43 & 50 & 54 & 55 \\
         & L2 prof. & 1 & 8 & 17 & 24 & 33 & 43 & 50 \\
         & & 2 & 18 & 37 & 47 & 54 & 56 & 58 \\
         & CEFR exp. & 1 & 7 & 20 & 32 & 35 & 44 & 49 \\
         & & 2 & 14 & 26 & 38 & 50 & 56 & 57 \\
         & Mixed & 1 & 11 & 22 & 36 & 45 & 52 & 57 \\
         & & 2 & 26 & 44 & 48 & 55 & 58 & 59 \\
         \midrule
         Adverbs & L2 sp. & 1 & 4 & 16 & 27 & 31 & 37 & 40 \\
         & & 2 & 12 & 26 & 33 & 41 & 50 & 53 \\
         & & 3 & 15 & 36 & 46 & 53 & 54 & 55 \\
         & L2 prof. & 1 & 8 & 17 & 29 & 36 & 41 & 46 \\
         & & 2 & 11 & 30 & 41 & 48 & 51 & 55 \\
         & CEFR exp. & 1 & 14 & 31 & 44 & 48 & 51 & 54 \\
         & & 2 & 13 & 33 & 44 & 51 & 56 & 58 \\
         & Mixed & 1 & 14 & 32 & 44 & 49 & 49 & 54 \\
         & & 2 & 25 & 48 & 54 & 59 & 60 & 60 \\
         \bottomrule
    \end{tabularx}
    \caption{Effect of different $d$ values}
    \label{tab:results:linear:scale:d}
\end{table*}

A more qualitative analysis reveals that 
for group 1 (interjections etc.)
for non-experts
with one vote, the hardest and easiest item is the same as with five votes, whereas with two votes, the two easiest and the three hardest are the same as with five votes.
For CEFR experts
with one vote, the three easiest items are the same as with three votes, whereas with two votes, the two hardest items are also the same as with five votes.
For L2 professionals
with one vote, the easiest item is the same as with three votes whereas with two votes, also the two hardest items are the same as with three votes.
However, many of the rank differences are small, i.e. the two hardest items for group 1 (interjections etc) for L2 professionals with one vote are the reverse order of two and three votes.
If one were to start from a truly unlabeled set of items without indications of level, or the number of different levels present in the data, one can only rely on relative ranks. These results indicate a certain stability when it comes to the extremes of the scale, i.e. which items are easiest and which items are hardest. 

In order to account for small differences in ranks, we also compute how many items are ``at the same rank" when counting as the same rank items within a difference of $d$, with $d$ varying from $1$ to $5$ (n.b. $d=0$ is equivalent to the same rank, column `Same' in Table \ref{tab:results:linear:scale:oop}; cf `Tolerance'). If we take as an example the ranking in Figure \ref{fig:oop:illustration}, at $d = 1$, one would count as being of equal rank the item $A$ (in addition to item $C$), as the rank difference is 1. At $d = 3$, one would also consider as being of equal rank items $B$ and $D$, as they are below or equal to $3$. 
Table \ref{tab:results:linear:scale:d} shows the results; we repeat $d=0$ for comparison purposes.

It can be said that the lists derived from a sub-sample of votes are different from the lists derived from all votes. However, when relaxing the notion of ``equivalence" as has been done by varying $d$, one can see that the difference is not as big as one might think at first. At $d=2$, which means deviations of two ranks (out of 60) or less are counted as equal, around 84\% of the lists are ``equal" to the lists derived from full votes for the aggregated versions (82\% for interjections, 80\% for verbs and 90\% for adverbs). Again, it can be observed that sampling over the whole crowd produces more stable results than sampling within a group. It can further be observed that the aggregated votes tend to be on par with expert judgments, if not surpassing them.

%% file: sections/6_6_time_invest.tex
\begin{table*}[htbp]
    \centering
    \begin{tabular}{lccccccccc}
    \toprule
         &  Group 1 & min & max & Group 2 & min & max & Group 3 & min & max \\
         \midrule
         L2 speakers & 36&3&164 & 38&6&260 & 44&3&227 \\
         L2 prof. & 41&13&43 & 26&14&44 & 24&14&44 \\
         CEFR exp. & 32&28&39 & 34&23&41 & 36&21&60 \\
         \midrule
         Average & 36&& & 32&& & 34&& \\
         \bottomrule
    \end{tabular}
    \caption{Average number of seconds per task and group }
    \label{tab:result:avg:time:cs}
\end{table*}

Table \ref{tab:result:avg:time:cs} shows the average time taken per crowd background and MWE group.
Despite the presence of outliers in the non-expert crowd data, 
crowdsourcing in a best-worst scaling scenario takes on average 30-40 seconds per task. To rank 60 items presented through 326 tasks with one vote would claim $\approx$~2,5-3 hours. Rankings do not seem to change drastically after the first three votes are collected, so the minimal time investment for 3 votes are estimated to approximately 8-9 hours for one project. 

Table \ref{tab:result:time} shows the comparison between observed times in the crowdsourcing project and reported times for direct annotation by the CEFR experts. It should be noted that for expert direct annotation, the times indicated in Table \ref{tab:result:time} are approximated by dividing the reported time needed to finish all three lists by three. It should also be borne in mind that experts went through all 326 tasks per project. It can be observed that direct expert annotation claims 15-90 minutes per project. This is at least five times as fast as the crowdsourcing experiment.

\begin{table*}[htbp]
    \centering
    \begin{tabularx}{\textwidth}{XXXXXXX}
    \toprule
         &  \multicolumn{2}{l}{Group 1 (interjections)} & \multicolumn{2}{l}{Group 2 (verbs)} & \multicolumn{2}{l}{Group 3 (adverbs)} \\
         & CS & Direct & CS & Direct & CS & Direct \\
         \midrule
         Expert 1 & 217 & $\approx$~90 & 225 & $\approx$~90 & 148 & $\approx$~90 \\
         Expert 2 & 155 & $\approx$~15 & 129 & $\approx$~15 & 117 & $\approx$~15 \\
         Expert 3 & 156 & $\approx$~20 & 199 & $\approx$~20 & 327 & $\approx$~20 \\
         \bottomrule
    \end{tabularx}
    \caption{Observed (crowdsourcing, CS) and reported (direct) times for experts for the two modes of annotation, in minutes}
    \label{tab:result:time}
\end{table*}
However, reliability and consistency of a (direct) labeling depend to a larger extent upon what kind of ranking scale annotators are offered and what their backgrounds are, and the effects are difficult to account for \cite[cf][]{o1995weighing}. It is easy to fall victim to a flawed design, inexperienced annotators or face problems hiring annotators, and the cognitive load of such an exercise is higher than in a crowdsourcing set-up \cite[e.g.][]{lesterhuis2017comparative}.

\begin{figure*}
    \centering
    \begin{subfigure}[b]{0.5\textwidth}
    \resizebox{.95\textwidth}{!}{
    \begin{tikzpicture}
    \begin{axis}[legend pos=outer north east, xlabel={Number of items in experiment}, ylabel={Number of combinations}]
    \addplot [only marks, mark = *] table {\datatablecomb};
    \end{axis}
    \end{tikzpicture}
    }
    \subcaption{Number of combinations}
    \label{fig:time:invest:number:items:combinations}
    \end{subfigure}%
    \begin{subfigure}[b]{.5\textwidth}
    \resizebox{.95\textwidth}{!}{
    \begin{tikzpicture}
    \begin{axis}[legend pos=outer north east, xlabel={Number of items in experiment}, ylabel={Time per person (minutes)}]
    \addplot [only marks, mark = *] table {\datatabletime};
    \end{axis}
    \end{tikzpicture}
    }
    \subcaption{Time per person}
    \label{fig:time:invest:number:items:time}
    \end{subfigure}
    \caption{Number of combinations and time per person with varying number of items}
    \label{fig:time:invest:numer:items}
\end{figure*}
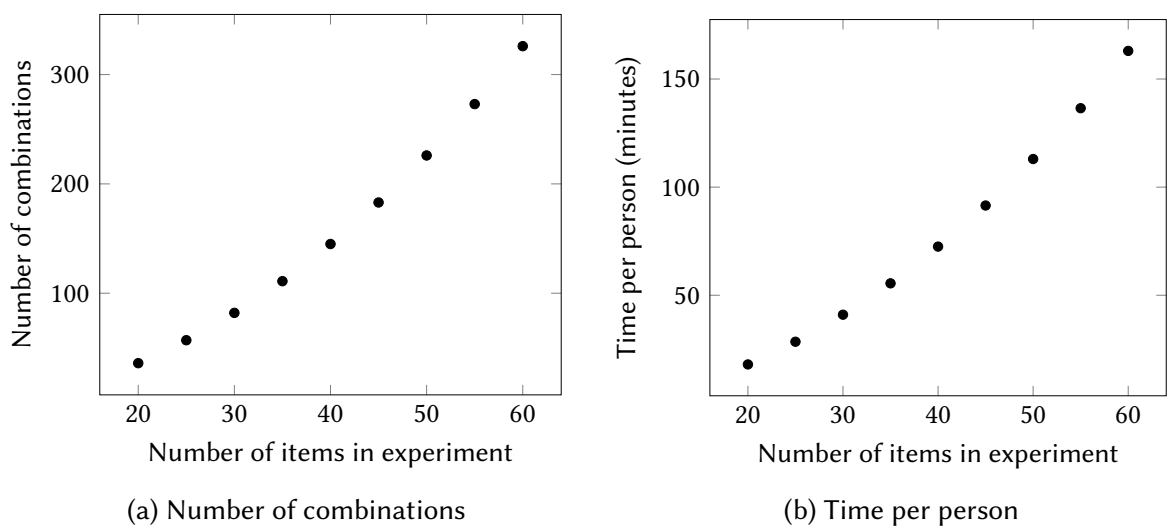

The time required to complete such a crowdsourcing experiment depends on the number of items that make up the experiment. Thus, for 20 items and 4 items per task, if one calculates with a mean response time of 30 seconds per task, it would take three crowdsourcers approximately 18 minutes each if one were to collect three votes per task.\footnote{If one wants to collect three votes per task, the minimum required number of participants is three, as no (registered) participant will be shown the same task twice.}
Figure \ref{fig:time:invest:number:items:combinations} shows the number of combinations in the experiment when varying the number of items from 20 to 60 in increments of 5.
Figure \ref{fig:time:invest:number:items:time} shows the amount of time it would take each person on average to complete the project under the above constraints. 
It can be noted that there seems to be a curvilinear relationship between the number of items and the number of combinations; this relationship would be exponential if it were not for the redundancy-reducing algorithm used. 
If one looks at the time per person in relation to the number of items, the relation seems to mimic the relation between number of items and number of combinations. Further, doubling the number of crowdsourcers (from 3 to 6) leads to a reduction of time per crowdsourcer by half: for 20 items and 4 items per task with a mean response time of 30 seconds per task, it would take six crowdsourcers approximately 9 minutes each if one were to collect three votes per task.

\pgfplotstableread{
X Y
20 36
25 57
30 82
35 111
40 145
45 183
50 226
55 273
60 326
}\datatablecomb

\pgfplotstableread{
X Y
20 18
25 28.5
30 41
35 55.5
40 72.5
45 91.5
50 113
55 136.5
60 163
}\datatabletime

%% file: sections/7_discussion.tex
Among the burning questions in emerging crowdsourcing projects -- within the domain of language learning -- the three methodological questions below remain the most important at the current stage of development:
\begin{enumerate}
\item Who can be the crowd -- with regards to the \textit{background} of crowdsourcers? 
\item How can 
\textit{reliable} annotations be achieved with regards to design, number of answers and number of contributors? and
\item How should the \textit{the results be interpreted} with regards to both research and practice?
\end{enumerate}

The biggest gap that we have tried to fill with this study concerns the first (1) question, i.e. whether crowdsourcing as a method in language learning -- within a limited domain of L2 resource annotation -- could be used without explicit control for the background of the crowd. 

Our results convincingly show that non-experts can perform on par with experts. We have seen that crowds with different backgrounds agree very well with each other, in comparison to previous research where CEFR raters of essays have often reached fairly low agreement. 
In fact, a mixed background crowd reaches ``average" rankings faster. Note here that these conclusions are true of annotation carried out in a \textit{comparative judgment} or \textit{best-worst scaling} setting 
whereas previous work on essay rating has been done based on scales such as the CEFR-scale similar to our direct-labeling experiment with the CEFR-experts.
To further confirm our findings, similar experiments need to be repeated for other languages, for other types of problems (e.g. annotation of texts for difficulty/readability), and for other sub-problems of a given problem (e.g. annotation of single vocabulary items for difficulty). 
Similar conclusions have been made in projects within citizen science, among others by 
\newcite{kullenberg2016citizen} 
where the experimental setup was not necessarily ``comparative" in character. This leaves room for further experimentation.  

In relation to question (2) the \textit{reliability} of annotations, we have seen how the design of an annotation task influences the results. Clearly, 
a more traditional method of annotation -- using expert judgments -- compares negatively to crowdsourced comparative judgments/best-worst scaling rankings. We have seen that experts do not agree with themselves when using comparative judgments versus categorical judgments, whereas the comparative judgment setting leads to homogeneous results between all groups of crowdsourcers regardless of their background, as shown in Section \ref{section:related:work}. 
According to \newcite{hovy2010towards} reliability of annotation of language resources has two types of major consequences, namely theoretical ones for shaping, extending and re-defining theories,
and practical ones for use in the classrooms, but also in teaching and assessment practices. Unreliably annotated data can lead to biased -- if not erroneous -- theoretical conclusions and generalizations, as well as influence teaching and assessing practices in unwanted ways, as discussed among others in \newcite{carlsen2012proficiency}. 

The above-stated \textit{theory--practice} dichotomy can be traced down to the proficiency dimension of the MWE items in our experiment. On the one hand, the ranked list represents Multi-Word Expressions according to their difficulty from the learner’s point of view and can thus be assumed to reflect stepwise development of their phraseological competence, which is of immediate interest to theoretical studies on L2 development. On the other hand, the scale represents perceptions of L2 professionals and -- hypothetically -- reflects their reasoning about what to teach/assess and in which order to do so based on their practical experience from teaching and assessing language learners, and has an immediate relevance for practical applications in ``real life", including use in automatic solutions for language learning. It is very encouraging to see that the two perspectives (theoretical-developmental and practical) produce similar results and are so much in agreement with each other. However, this harmony can be observed only as long as we view vocabulary development as a continuum as opposed to groups of items belonging to 
one of a number of categorical proficiency levels. 

In fact, both dimensions -- theoretical and practical -- are equally important. To understand how to teach and what to teach (practical dimension), we need to understand how learning is happening and (among others) observe which linguistic and cognitive aspects develop and in which order. While the produced scales give us material to study development of phraseology from a theoretical point of view, it is not obvious how to apply these scales to practical use (question (3) above) in teaching, assessment and Intelligent CALL, where categorical representations of proficiency are more customary and readily applicable. There are no indications in our crowdsourcing results
as to where to draw the line between one level of proficiency and the other. We are not unique in facing these troubles, even though in other areas it can be a vice versa case: 

\begin{quote}
A weakness in this line of work is that SLA researchers have most often chosen to treat proficiency as a categorical variable and then have assessed mean differences in complexity values across proficiency groupings. Yet, this practice of converting interval variables (i.e. individual proficiency scores of some kind) into categorical ones (i.e. participants grouped by nominal proficiency levels) has always been criticized by statisticians because it discards much useful information. More specifically, it does away with the variance of continuous scores and leads to unreliability and increased likelihood of Type II errors \cite[e.g.][]{troncoso2010statistical}, that is, the problem of failing to detect a difference, relationship, or effect that is in fact present because of some psychometric methodological problem, such as lack of power or (in the case at hand) lack of variance in the observations. It would be profitable in future work, therefore, to accumulate evidence from designs where both complexity and proficiency are treated as interval scales.
\begin{flushright}
\cite[p.~131]{ortega2012interlanguage}
\end{flushright}
\end{quote}

This is a current challenge that needs to be addressed in the future 
\cite[e.g.][]{paquot2020using}. Proficiency levels are always rather arbitrary \cite{hulstijn2010developmental}  as is also noted by the authors of CEFR \cite{councilofeurope2001} who caution that ``any attempt to establish `levels' of proficiency is to some extent arbitrary, as it is in any area of knowledge or skill. However, for practical purposes it is useful to set up a scale of defined levels to segment the learning process for the purposes of curriculum design, qualifying examinations, etc." (p.~17). 
To summarize this part of the discussion, we view our results as a strong argument for treating vocabulary development as a continuum, while we also recognize the need to establish ways to partition vocabulary by levels of proficiency where these items can be taught. 

On the practical side of crowdsourcing, our results show that a good and reliable agreement within a mixed crowd can be reached with two to three votes per task by at least three different voters. Considering these results, it might be interesting to use the same methodology for essay grading, especially since results from various experiments which have looked at inter-rater agreement in marking essays according to categorical proficiency levels have been less promising \cite[cf][]{carlsen2012proficiency,diez2012use}. 


One of the limitations  of the current setup lies in the use of the combinatorial algorithm which we apply to calculate the task pairings. As stated, we only achieve 77\% \emph{non-redundant} combinations, which means that certain pairs of expressions are included more than once and thus get more votes than other combinations, which might skew the picture.
More involved statistical methods such as balanced incomplete block design (BIBD) \cite{yates1936incomplete} can be used to circumvent this problem. However, such methods impose hard constraints on the number of items and the number of items per task and not all combinations of number of items and items per task are able to satisfy these constraints. To the best of our knowledge, there exists no solution to the BIBD constraints for 60 items with 4 items per task.

The two methods of annotation -- crowdsourcing by unknown crowd versus annotation by approved experts -- have different dimensions of pros and cons. Here we have seen that time versus reliability can outweigh each other. In addition, one needs to consider that when using crowdsourcing, one has little control over the participant group and the time. Hence, neither method is superior on all accounts, but both are appropriate as long as one is aware of their weaknesses and strengths. 
If one is able to pay CEFR experts, one may get faster results. However, as seen in this study, one would need a large number of experts to reach consensus. Thus, expert knowledge can be fast and reliable \emph{if} a large enough number of experts is consulted, to counteract the bias of 
individual subjective opinions, but it is also expensive.
If one does not have access to experts for various reasons, one can use crowdsourcing as an alternative to derive a relative ranking of expressions. The resulting ranking is similar regardless of whether one uses non-experts or experts, thus one may be able to realize such an experiment with non-experts only. In contrast to using experts for direct annotation, crowdsourcing is cheap however it takes longer time, both regarding the implementation and the actual crowdsourcing phase. Furthermore, with the set up we have chosen, one does not get concrete CEFR levels but rather a relative ranking.
This data can, however, potentially be partitioned into more or less discrete proficiency levels by various techniques, should one desire to do so. The exploration and experimentation in this direction is future work.

%% file: sections/8_conclusion.tex
In this study, we asked whether it influences the results in a crowdsourcing experiment aimed at ranking MWEs by difficulty if crowdsourcers are experts (L2 Swedish professionals) or non-experts (L2 Swedish learners / speakers). We set up different crowdsourcing experiments for the different target groups so as to be able to compare the results of different groups.
The presented experiment suggests that it does not matter for this type of experiment if the crowdsourcer is an L2 speaker or an L2 professional, as the results produced by L2 speakers of Swedish, teachers of Swedish and CEFR experts are highly correlated. Concerning the design of the annotation task, we have convincingly shown that comparative design is a winner in contrast to explicit labeling: one does not need to have recourse to expert knowledge, and the results are much more homogeneous.


Furthermore, we explored how the number of votes influences the results and we found that with only two votes, the difference in results on a scale 1-60 is insignificant in comparison to three votes. 
Additionally, we found that sampling from a mixed-background group tends to produce more stable results. Indeed, using a mixed crowd produces results similar to results obtained from only expert annotations.
This finding can further speed up crowdsourcing projects, since one can gather data with only one experiment instead of having to set up three distinct experiments for each target background.
We also found that L2 \emph{proficiency}, as measured by L2 professionals, does seem to correlate with L2 \emph{development}, collected through intuitive judgments by L2 speakers.

These findings suggest that crowdsourcing might be a viable method to create a ranking of expressions by difficulty even in the absence of gold standard data. 
Our results suggest that there is a strong incentive in exploring crowdsourcing for other languages (if getting a scale is sufficient).
For any new language and new item combination, we would suggest that the best-worst method be applied. There are reasons to believe that having strong ``anchor words'' for levels, i.e. words for which one knows the level with reasonable certainty, among the data can help create clusters around those with suggestions where to draw the line between one level and another, if there is a need for the pedagogical, assessment, CALL or other uses. 

Future studies could investigate whether the same methodology produces the same results when applied to, for example, single word expressions or essays. Another direction for future research, as shortly mentioned above, might be how to partition an unordered, unlabeled set of expressions into different proficiency levels based, for example, on clustering results. This might be achieved by adding certain \emph{anchor} expressions to the experiment, i.e. expressions of which one knows with a sufficient degree of certainty their true label (i.e. target level). As a possible starting point, one could take the easiest and the hardest expressions overall from a ranking experiment such as the one presented, as the agreement at the extremes (very easy and very hard expressions) tends to be much higher than in the middle of the scale. Further, one might want to investigate how core and peripheral vocabulary can be identified based on 
different kinds of annotations. 




